\documentclass[conference]{IEEEtran}
\usepackage{times}
\usepackage[numbers]{natbib}
\usepackage{multicol}
\usepackage[bookmarks=true]{hyperref}
\usepackage{booktabs}
\usepackage{graphicx}
\usepackage{amsmath}
\usepackage{amssymb}
\usepackage[T1]{fontenc}

\begin{document}

\title{More Structure, Not More Capacity:\\
Object-Centric Representations for Visuomotor Imitation Learning}

\author{Yi Li$^{1}$,\quad Alexandre Chapin$^{2}$,\quad Liming Chen$^{2}$,\quad
Jan Peters$^{1}$,\quad Alap Kshirsagar$^{3}$ \\[2pt]
{\small $^{1}$TU Darmstadt \qquad $^{2}$École Centrale de Lyon \qquad $^{3}$IIT Delhi - Abu Dhabi}}

\maketitle

\begin{abstract}
Robotic manipulation policies rely on pre-trained vision models that give either
a global scene embedding or a dense patch grid. Both mix task-relevant and
task-irrelevant features. Object-centric slot representations are a structured
alternative: they group features into a few per-object slots. We test what this
structure buys on ManiSkill3 PickCube-v1, with a frozen encoder and a
held-out-seed evaluation. Holding the policy, goal token, rendering, and
calibration fixed and changing only the encoder, a frozen object-centric SPOT
representation (DINO ViT-B/16 + Slot Attention) reaches $55.0\!\pm\!2.9\%$
success, $22.4\%$ above a dense DINO global-feature baseline
($32.6\!\pm\!1.5\%$), with the same trainable policy and no encoder
fine-tuning. More tokens alone do not help: a dense patch grid with $16\times$
the tokens performs no better than the global feature. Adding an explicit 2D
spatial goal and native-resolution rendering raises the full system to
$68.7\!\pm\!4.2\%$, just below a privileged 3D-oracle upper bound
($71.7\!\pm\!4.1\%$). An automated kinematic failure taxonomy then separates
spatial-precision (Near-Miss) failures from object-tracking (No-Grasp) failures:
spatial grounding reduces Near-Miss while leaving No-Grasp unchanged. The same
taxonomy transfers to the harder StackCube-v1 and points to occlusion as the
main bottleneck.
\end{abstract}

\IEEEpeerreviewmaketitle

\section{Introduction}
\label{sec:intro}

A visuomotor policy never sees the raw scene. It only sees what its visual
encoder keeps. The choice of representation therefore decides what information
reaches the policy. In robotics this question is most developed for mapping and localization, where
geometric, probabilistic, dense, and implicit map representations each shape
behavior differently~\cite{cadena_slam_2016}. It applies just as much to
manipulation: how does the structure of the visual representation affect
behavior when the policy must act on object and goal positions it never saw
during training? We answer this with a controlled, task-level study on a
simulated pick-and-place task.

Self-supervised vision transformers such as DINO~\cite{caron_emerging_2021}
transfer well to recognition. It is less clear how they behave when a policy
must act on scene configurations not seen in training. Recent work across image
classification~\cite{shi_last-vit_2026}, 3D perception~\cite{zhang_utonia_2026},
and gaze estimation~\cite{qin_unigaze_2025} points to one shared pattern: a
backbone trained under global or invariance-oriented supervision finds a
low-cost cue that holds in-distribution and breaks under shift, and in each case
the fix is structural rather than adding parameters.

This paper studies a similar pattern in visuomotor imitation. Under matched
conditions, we compare dense DINO features (both the global \texttt{[CLS]} token
and dense patch grids) against an object-centric slot representation built on
the same frozen backbone. We treat the slot representation as a small,
competition-based bottleneck. Slot
Attention~\cite{locatello_object-centric_2020}, as implemented in
SPOT~\cite{kakogeorgiou_spot_2024}, compresses the dense patches into a few
slots and makes each slot bind to one coherent region. This is an architectural
inductive bias toward object-level structure. Since the encoder stays frozen,
this bias has to come from the module placed on top of its features, not from
retraining the backbone. A strong pretrained backbone is not enough when the
task needs signals that the pretraining objective suppressed.

\noindent In this work, we make the following contributions:
\begin{enumerate}
  \item A systematic, matched-condition comparison of dense global, dense
        patch, and object-centric representations for visuomotor imitation,
        under a single policy and a held-out-seed protocol that measures
        generalization to novel object and goal placements (a task-level
        evaluation of representation choice).
  \item The finding that representational \emph{structure}, not capacity,
        drives generalization: object-centric slots beat dense features by
        $22.4\%$ under matched conditions, while a 392-token dense grid
        collapses, and neither bidirectional cross-attention nor a longer
        temporal window improves held-out success over their simpler counterparts.
        We read this as a similar shortcut-learning pattern to that seen in
        recent foundation-model work, and support that reading with what we can
        measure directly: how the slots bind, and how the failures are
        distributed.
  \item An automated, encoder-agnostic kinematic failure taxonomy that separates
        spatial-precision (Near Miss) from object-tracking (No-Grasp) failures, giving a
        task-level read of the trade-offs and failure modes of each representation and
        grounding choice. It transfers across tasks: applied unchanged to StackCube-v1,
        it points to occlusion as the main bottleneck.
\end{enumerate}

\section{Related Work}
\label{sec:related}

\paragraph{Object-centric representations.}
Slot Attention~\cite{locatello_object-centric_2020} routes image patches into
discrete, permutation-equivariant slots through iterative competition.
Subsequent work extended this to video~\cite{kipf_conditional_2022}, to
feature-space reconstruction on frozen backbones (DINOSAUR~\cite{seitzer_bridging_2023}),
and to diffusion decoders (SlotDiffusion~\cite{wu_slotdiffusion_2023}). We build
on SPOT~\cite{kakogeorgiou_spot_2024}, which stabilizes slot binding via
autoregressive patch-order permutations and self-training through attention-mask
distillation on a frozen DINO ViT-B/16 backbone. We choose it over these
alternatives because it gives stable slot binding on real images within an
11\,GB VRAM budget and runs as a feed-forward encoder with no decoder at
inference, which suits our frozen-feature, single-GPU setting. We use SPOT as a
fixed feature extractor and ask what its structure gives the policy. Most
directly related to our setting, \citet{chapin_spotlighting_2026} present a
large-scale comparison of global, dense, and object-centric representations and
show, with frozen encoders, that object-centric slots improve manipulation
generalization under visual distribution shifts (lighting, texture, clutter). We
build on this in a complementary direction. Instead of visual shifts, we test
generalization to unseen object and goal placements on a single task under a
held-out-seed protocol; we isolate representational structure from token count;
we add an explicit spatial-grounding study; and we introduce a kinematic failure
taxonomy that applies unchanged across tasks.

\paragraph{Shortcut learning in vision foundation models.}
\citet{shi_last-vit_2026} identify lazy aggregation in ViTs: the \texttt{[CLS]}
token relies on low-frequency background content and generalizes poorly when
foreground--background correlation shifts. \citet{zhang_utonia_2026} show that
cross-domain 3D encoders exploit scan patterns and gravity bias as domain
identifiers, removable only with targeted structural interventions.
Work on gaze estimation~\cite{qin_unigaze_2025} shows that invariance-oriented
pretraining discards the fine-grained geometric signal gaze estimation
requires, and that a domain-matched, detail-preserving objective recovers it.
The same pattern appears in robotic manipulation:
\citet{chapin_spotlighting_2026} show that global and dense features entangle
task-relevant and irrelevant cues and lose robustness under scene changes. These
works span 2D recognition, 3D perception, fine-grained regression, and
manipulation; we add a controlled visuomotor-imitation study and, as in those
settings, find the effective remedy to be structural rather than parametric.

\paragraph{Visuomotor imitation with frozen features.}
R3M~\cite{nair_r3m_2022} and MVP~\cite{radosavovic_real-world_2022} show that
frozen self-supervised features support manipulation without end-to-end
fine-tuning. ACT~\cite{zhao_learning_2023} and Diffusion
Policy~\cite{chi_diffusion_2023} establish action chunking and sequence-based
policy decoders as strong behavior-cloning baselines. We build on this
paradigm and isolate the \emph{encoder structure} as the critical variable,
holding the policy and the frozen-feature setting fixed.

\section{Method}
\label{sec:method}

\begin{figure*}[t]
  \centering
  \includegraphics[width=\linewidth]{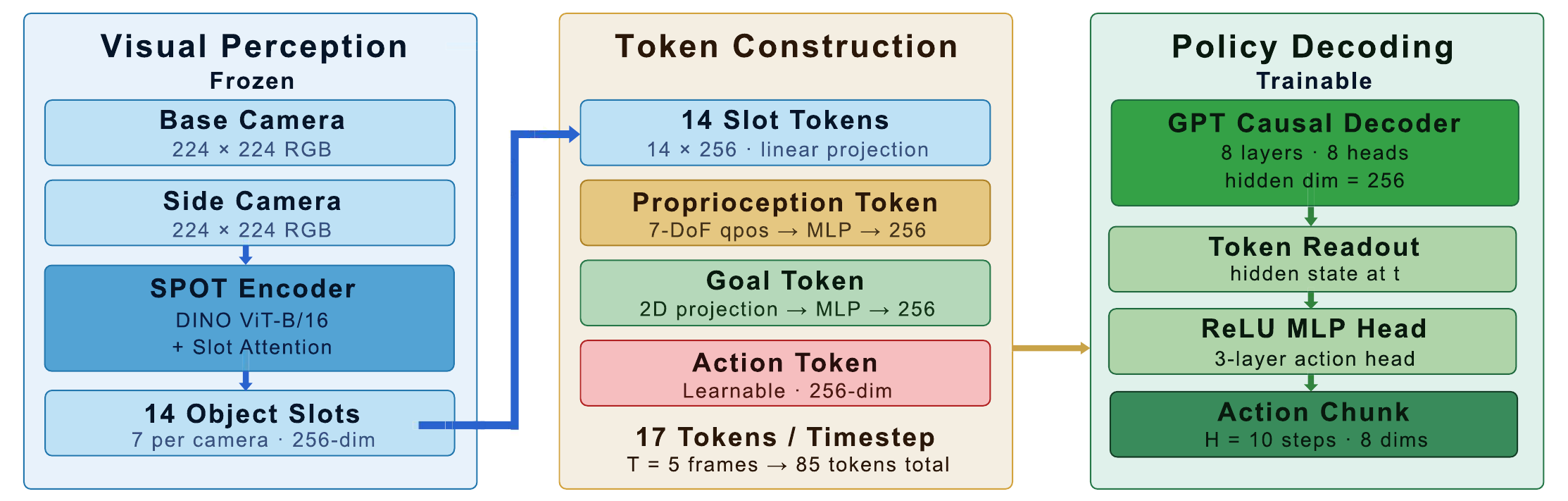}
  \caption{System overview. RGB images from two fixed cameras are encoded by
    the frozen SPOT encoder (DINO ViT-B/16 + Slot Attention) into 7 object
    slots each. The 14 visual slots, 1 proprioception token, 1 goal token, and
    1 learnable action token form a 17-token sequence per timestep. A causal Transformer policy
    predicts a 10-step action chunk.}
  \label{fig:arch}
\end{figure*}

Figure~\ref{fig:arch} shows the pipeline: a frozen visual encoder, a token
construction stage, and a trainable autoregressive policy.

\subsection{Frozen SPOT Encoder}
\label{sec:encoder}
Two RGB cameras (a base and a diagonal-side view, both $224\!\times\!224$)
observe the scene from complementary viewpoints. Each image is processed
independently by the frozen SPOT encoder, which chains a DINO ViT-B/16 backbone
with a Slot Attention module to compress $196$ patch tokens into $K\!=\!7$
object slots $S\in\mathbb{R}^{7\times256}$. K=7 is fixed by the publicly released SPOT 
checkpoint and is not tuned; we inherit it as-is. Slot
Attention~\cite{locatello_object-centric_2020} normalizes attention over the
\emph{slot} dimension, so each patch activates at most one slot strongly:
\begin{gather}
  M_{i,j}=\frac{k(X_i)\cdot q(S_j)}{\sqrt{D}}, \quad W_{i,j}=\frac{\exp(M_{i,j})}{\sum_{l=1}^{K}\exp(M_{i,l})}, \\
  S^{(t)}=\mathrm{GRU}\left(S^{(t-1)},\,W^\top v(X)\right)
\end{gather}
This competition forces object-level binding and keeps the representation from
collapsing to scene-level statistics. Slots from both cameras are
concatenated into $S_{\mathrm{vis}}\!\in\!\mathbb{R}^{14\times256}$. We use the
publicly released SPOT checkpoint, pretrained on COCO and kept frozen throughout
all experiments; observed performance differences therefore reflect properties
of the representation, not task-specific fine-tuning.

\subsection{Token Construction and Spatial Grounding}
\label{sec:tokens}
At each timestep, four token types are projected to a shared 256-dimensional
space and concatenated into a 17-token sequence: (i)~14 visual slot tokens;
(ii)~1 proprioception token (7-DoF joint positions via MLP); (iii)~1 goal
token (below); (iv)~1 learnable action token queried at readout. We study three
goal-conditioning conditions to isolate the contribution of spatial
information:
\begin{itemize}
  \item \textbf{Pure Visual:} no goal token; the policy relies on visual slots
        only.
  \item \textbf{2D Spatial Projection:} the 3D goal is projected onto both
        camera planes via intrinsics/extrinsics, giving a 4D pixel vector
        $g=[u_b,v_b,u_s,v_s]$ normalized to $[0,1]$ and encoded by a 3-layer
        MLP.
  \item \textbf{3D Oracle:} ground-truth 3D world coordinates from the
        simulator. This is a privileged upper bound and is not deployable.
\end{itemize}
Both the 2D projection and the 3D oracle use the simulator's ground-truth goal
location; they differ only in how the goal is represented (pixel vs.\ world
coordinates), not in the information they assume. The 2D projection is therefore
not self-supervised; deriving the goal anchor from observation alone (e.g.\ from
the slot masks) is future work.

\subsection{Autoregressive Sequence Policy}
\label{sec:policy}
Our policy is a causal Transformer decoder trained for behavior cloning. We
reuse the GPT (minGPT-style) decoder from the VQ-BeT implementation in
LeRobot~\cite{lee_vqbet_2024}; unlike VQ-BeT, we predict continuous actions
directly, with no vector quantization. An 8-layer, 8-head causal Transformer policy with hidden dimension
256 processes a temporal window of $T\!=\!5$ frames ($85$ tokens total) under a
lower-triangular causal mask. The hidden state of the action token at the
current timestep passes through a 3-layer ReLU MLP to predict a 10-step action
chunk ($H\!=\!10$, action dimension 8). The 7 continuous joint dimensions are
Z-score normalized; the gripper dimension bypasses normalization and is decoded
by a hard step function.

\subsection{Representation Baselines}
\label{sec:baselines}
Under identical policy architecture and goal conditioning, we compare:
\begin{itemize}
  \item \textbf{DINO Global [CLS]:} 1 token per camera (2 total).
  \item \textbf{DINO 4$\times$4 patches:} 16 tokens per camera (32 total),
        spatially subsampled.
  \item \textbf{DINO 14$\times$14 dense:} 196 tokens per camera (392 total);
        memory forces $T\!=\!1,\,H\!=\!1$.
  \item \textbf{Object-Centric (SPOT):} 7 slots per camera (14 total).
\end{itemize}
The dense $14\!\times\!14$ baseline is a deliberate stress test: it has the most
tokens but, forced to a single frame and single-step prediction by GPU memory,
it removes temporal context.

\section{Experimental Setup}
\label{sec:setup}

\paragraph{Task and simulator.}
All experiments use ManiSkill3~\cite{tao_maniskill3_2025} with the
deterministic \texttt{physx\_cpu} backend. The primary task,
\textbf{PickCube-v1}, uses a Franka Panda arm to grasp a red cube and place
it at a randomly initialized 3D target. We additionally train and evaluate the 
same architecture on \textbf{StackCube-v1} (stack a red cube on a green cube) to
characterize failure modes under higher occlusion. Training uses 1{,}000 expert
demonstrations from the simulator's motion planner, replayed at $100\%$ success.

\paragraph{Evaluation on unseen initializations.}
Object and goal positions are randomized at each episode. Training uses seeds
$0$--$9{,}999$ and evaluation uses seeds $\geq\!10{,}000$, so the policy is
evaluated only on placements it never saw during training (the seed ranges are
disjoint). This protocol measures generalization to novel placements rather than
memorization.

\paragraph{Evaluation protocol.}
Checkpoints are saved every 50 epochs and swept against 200 held-out episodes.
The best checkpoint undergoes three independent stability runs of 300 episodes
each; we report mean$\,\pm\,$std. Episodes terminating in fewer than 10 steps
are discarded as trivial successes. Because the encoder is frozen and only a lightweight
module is trained on top of its features, this setup is a \emph{decoder probing}
of frozen representations: it measures how much task-relevant signal a moderate
non-linear decoder can extract, which is closer to the deployment regime than
either linear probing or full fine-tuning.

\paragraph{Failure taxonomy.}
Failed episodes from a separate 200-episode run are classified by an automated
kinematic script (inputs: action variance, the \texttt{is\_grasped} flag, and
tool center point (TCP)-to-goal distance) into five mutually exclusive
categories applied in priority order: \textbf{Idle} (mean action variance below
threshold), \textbf{No Grasp} (\texttt{is\_grasped} never true), \textbf{Drop}
(grasped but released before the end), \textbf{Near Miss} (minimum TCP-to-goal
distance $<\!0.1$\,m but final placement failed), and \textbf{Unknown}. The
script uses no visual data and is encoder-agnostic.

\section{Results and Analysis}
\label{sec:results}

\begin{table}[t]
\centering
\caption{Success rates (SR) on PickCube-v1 (held-out seeds $\geq\!10{,}000$,
  mean$\,\pm\,$std over $3\!\times\!300$ episodes). The representation
  comparison (\emph{upper block}) holds policy, goal token, rendering
  resolution ($128^2$), and approximate calibration constant, so that only the
  visual encoder changes. The spatial-grounding and resolution ablation
  (\emph{lower block}) uses SPOT throughout. $\dagger$~forced to
  $T\!=\!1$, $H\!=\!1$ by GPU memory. $\ddagger$~privileged simulator state;
  oracle upper bound only.}
\label{tab:main}
\begin{tabular}{lcc}
\toprule
\textbf{Configuration} & \textbf{Tokens} & \textbf{SR (\%)} \\
\midrule
\multicolumn{3}{l}{\textit{Representation comparison ($128^2$, approx.\ cal., 2D goal)}} \\
\midrule
DINO 14$\times$14 dense\,$^\dagger$ & 392 & $1.0$ \\
DINO Global [CLS]                   & 2   & $32.6\pm1.5$ \\
DINO 4$\times$4 patches             & 32  & $31.7\pm3.0$ \\
Object-Centric (SPOT)               & 14  & $55.0\pm2.9$ \\
\midrule
\multicolumn{3}{l}{\textit{Spatial grounding and resolution ablation (SPOT)}} \\
\midrule
Pure visual (no goal)               & 14  & $31.0\pm2.8$ \\
+ 2D goal, approx.\ cal., $128^2$   & 14  & $55.0\pm2.9$ \\
+ 2D goal, exact cal., $128^2$      & 14  & $58.7\pm3.4$ \\
\textbf{+ 2D goal, exact cal., $224^2$ (full system)} & \textbf{14} & $\mathbf{68.7\pm4.2}$ \\
3D Oracle\,$^\ddagger$              & 14  & $71.7\pm4.1$ \\
\bottomrule
\end{tabular}
\end{table}

\subsection{Object-Centric vs.\ Dense Representations}
\label{sec:res-repr}
\begin{figure}[t]
  \centering
  \includegraphics[width=\linewidth]{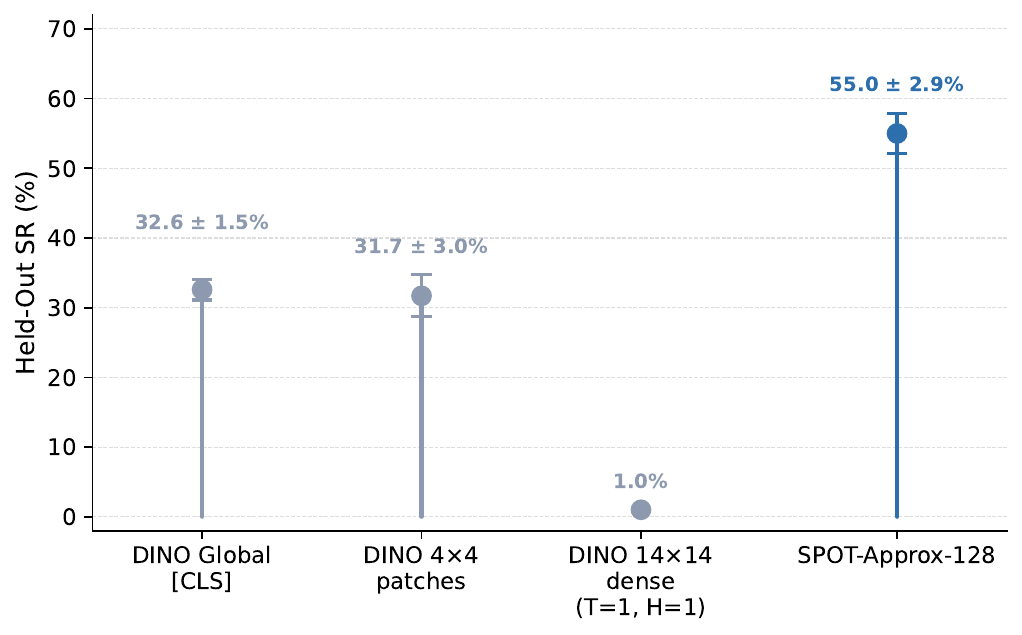}
  \caption{Success rate across representation types under matched conditions; the
  object-centric bar (shown as SPOT-Approx-128) is the SPOT representation at
  $128^2$ with approximate calibration, i.e.\ the Object-Centric (SPOT) row of
  Table~\ref{tab:main}. Object-centric grouping matters; token count does not.}
  \label{fig:results_main}
\end{figure}

The upper block of Table~\ref{tab:main} shows the matched comparison. All four
configurations use the same policy, the same 2D goal token, $128^2$ rendering,
and approximate calibration; only the encoder changes. The object-centric SPOT
representation reaches $55.0\!\pm\!2.9\%$, against $32.6\!\pm\!1.5\%$ for DINO
Global [CLS] and $31.7\!\pm\!3.0\%$ for DINO $4\!\times\!4$ patches. That is a
$22.4\%$ gap from a change in structure alone, with the same trainable policy
and no encoder fine-tuning. Token count does not track success: the
$4\!\times\!4$ grid carries $16\times$ the tokens of the global feature but
performs the same ($31.7$ vs $32.6$), and the object-centric representation
beats it with fewer tokens. The dense $14\!\times\!14$ baseline does collapse to
$1.0\%$, but only after GPU memory forces it to $T\!=\!1,\,H\!=\!1$ and strips
temporal context, so we do not read that as a token-count effect. \textbf{The
gain comes from object-level structure, not token count}
(Fig.~\ref{fig:results_main}).

\subsection{Why It Works: A Capacity-Limited Bottleneck}
\label{sec:res-why}
\begin{figure}[t]
  \centering
  \includegraphics[width=\linewidth]{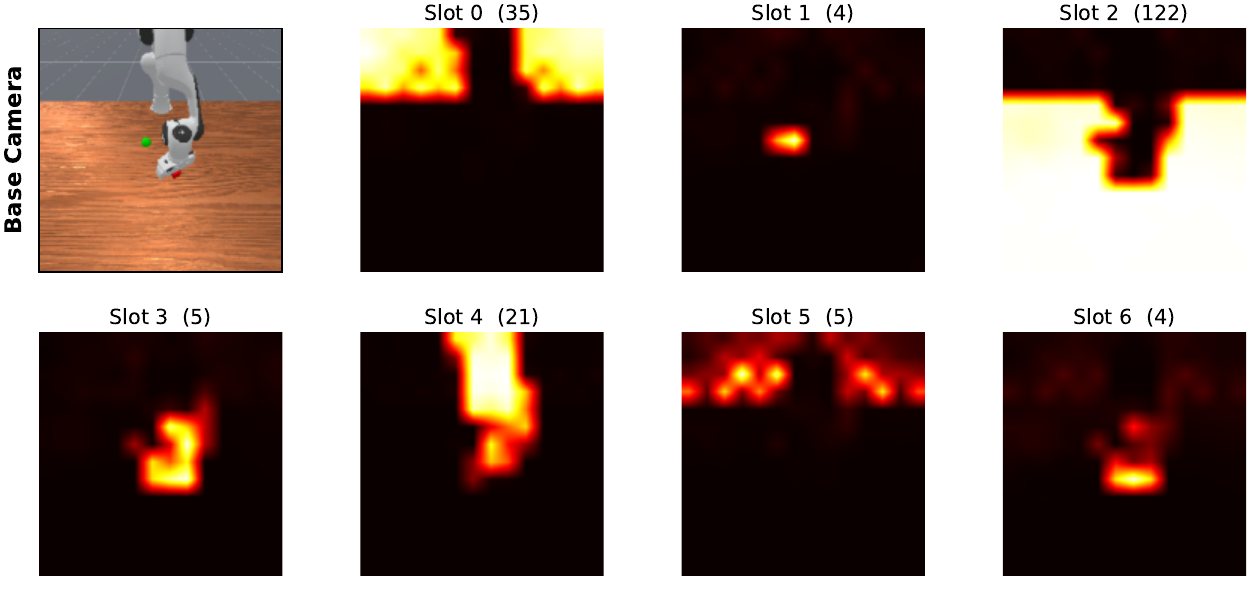}
  \caption{Per-episode slot masks from the base camera. A subset of slots binds
    to the cube, goal marker, and arm, while the rest cover the background.
    Numbers in parentheses give the patches (of 196) assigned to each slot; the
    largest slot (122) covers the table background, while smaller slots bind
    precisely to the cube and goal. This competition-based grouping is the
    structural bottleneck the dense baselines lack.}
  \label{fig:slot_viz}
\end{figure}

We read this result through what the architecture does. Slot Attention
compresses 196 patches per view into seven slots. The softmax competition over
the slot dimension makes each patch commit to a single slot, so no slot can take
in the whole image at once. Scene-level statistics are therefore not available
as a shortcut, and the limited capacity goes to coherent object regions instead.
Figure~\ref{fig:slot_viz} shows this: in clean episodes, some slots bind to the
cube, the goal marker, and the arm, while the rest cover the background. The
per-slot patch counts make the competition explicit: one dominant slot absorbs
most of the 196 patches (the table background), while smaller slots bind
precisely to the cube and goal. This is
the manipulation version of the pattern in Section~\ref{sec:intro}. A dense
global feature is free to mix information across the whole image. It keeps
co-occurrence cues from the training set, such as object textures, goal
positions, and background statistics that vary together during training, and it
breaks when a new initialization breaks those cues. We read the gap as
consistent with the lazy-aggregation account of~\citet{shi_last-vit_2026}. We do
not measure \texttt{[CLS]} attention directly. What we report instead is the
slot binding above and the spatial-precision failure profile in
Section~\ref{sec:res-grounding}.

\subsection{Fusion and Temporal Capacity Do Not Transfer}
\label{sec:res-fusion}

\begin{table}[t]
\centering
\caption{Multi-view fusion without explicit goal conditioning, on PickCube-v1.
  Training success rates are approximate, read off during training on seen
  initializations; held-out success rates are over $3\!\times\!300$
  held-out-seed episodes. Cross-attention looks far better during training but
  matches concatenation on held-out seeds, so its extra capacity does not
  transfer.}
\label{tab:fusion}
\begin{tabular}{lcc}
\toprule
\textbf{Fusion strategy} & \textbf{Train.\ SR (\%)} & \textbf{Held-out SR (\%)} \\
\midrule
Token concatenation           & $\approx38.6$ & $31.0$ \\
Bidirectional cross-attention & $\approx60.0$ & $31.0$ \\
\bottomrule
\end{tabular}
\end{table}

We next ask whether adding \emph{parameters} can recover what structure gives
for free. Without a goal token, token concatenation and bidirectional
cross-attention reach the same held-out SR of $31.0\%$ (Table~\ref{tab:fusion}),
measured over three runs of 300 episodes. Their training-time behavior differs
sharply: cross-attention reaches about $60\%$ on seen initializations against
roughly $38.6\%$ for concatenation. We treat these training-time numbers only as
a qualitative signal, since they are read off during training on seen seeds
rather than under the held-out protocol; the point is the direction, not the
level. The extra parameters of cross-attention fit the seen placements but do
not transfer to held-out ones. Concatenation has no such dedicated capacity; it
leaves cross-view integration to the causal self-attention over the 17-token
sequence. \textbf{We therefore prefer token concatenation to a parametric fusion
module.} One caveat on scope: this comparison was run without a goal token,
where success sits at its floor for both strategies and the real limit is the
missing spatial anchor.

A second capacity axis points the same way. Here we keep an early, pure-visual
configuration fixed and change only the temporal window. Success rate does not
move with history length ($6.7\%$ at $T\!\in\!\{5,20,40\}$), while a shorter
window trains about $4.5\times$ more epochs in the same compute budget. The low
absolute value here is from before the spatial-grounding and resolution fixes;
the point is that success does not change with $T$, not its level. Taken
together: without an explicit spatial anchor, neither more fusion parameters nor
more history turns into better generalization.

\subsection{Spatial Grounding and the Failure Profile}
\label{sec:res-grounding}
The lower block of Table~\ref{tab:main} isolates spatial grounding and
rendering resolution, with the SPOT encoder fixed. Pure-visual success rate is
$31.0\!\pm\!2.8\%$. Adding a 2D goal projection (approximate calibration,
$128^2$) raises it to $55.0\!\pm\!2.9\%$ ($+24\%$). Replacing approximate with
exact intrinsics and extrinsics adds $+3.7\%$, to $58.7\!\pm\!3.4\%$. So a
consistent but slightly biased anchor already recovers most of the gain;
pixel-perfect calibration helps but is not critical. Rendering natively at
$224^2$ removes an upscaling artifact that let the cube and goal share one pixel
cell at close placement distances. This adds a further $+10.0\%$, to
$68.7\!\pm\!4.2\%$, just below the 3D-oracle ceiling ($71.7\!\pm\!4.1\%$).

\begin{figure}[t]
  \centering
  \includegraphics[width=\linewidth]{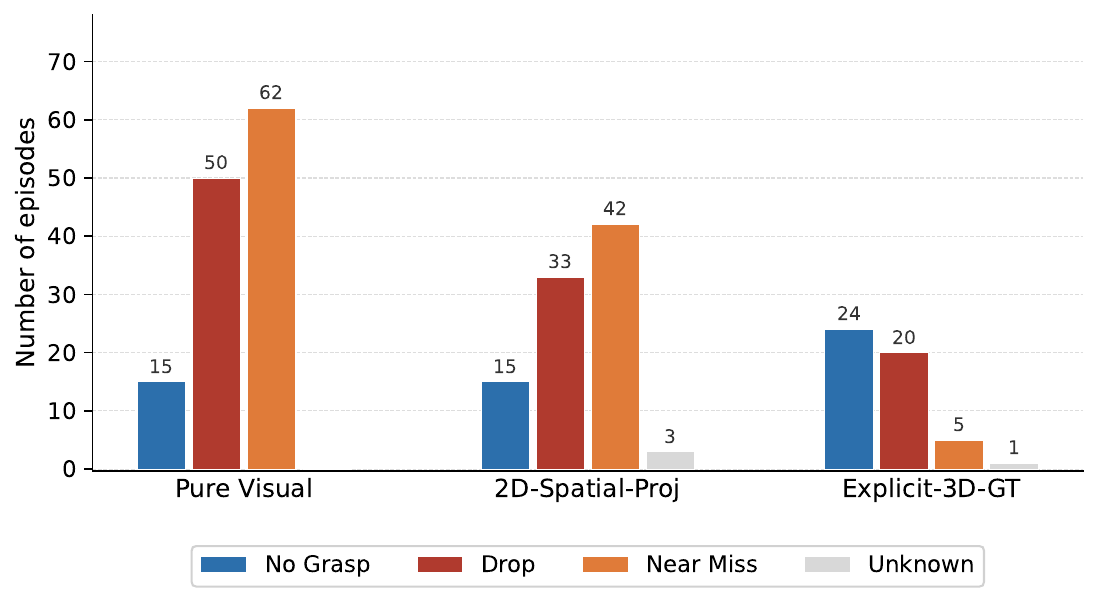}
  \caption{Failure breakdown across goal-conditioning conditions. Near-Miss
    (spatial precision) drops sharply with grounding; No-Grasp (object
    tracking) does not. The two bottlenecks are invisible to binary success
    rate.}
  \label{fig:taxonomy}
\end{figure}

The failure taxonomy (Fig.~\ref{fig:taxonomy}) makes this concrete, and shows
that binary success rate hides two separate bottlenecks. As goal conditioning
improves from pure visual to 3D oracle, Near-Miss failures (the robot reaches
the goal area but fails the final placement) fall from $31\%$ to $21\%$ to
$2.5\%$. No-Grasp failures are \emph{not} reduced by spatial grounding (15, 15,
and 24 episodes per 200 across the three conditions). The two modes need
different fixes. Near Miss is fixed by spatial grounding. No-Grasp needs better
occlusion handling or temporal slot consistency (e.g.\ SAVi++~\cite{elsayed_savi_2022}).
The taxonomy therefore separates spatial-precision failures from object-tracking
failures, which makes it possible to fix the right one rather than tune
blindly. Beyond aggregate success rate, this gives a task-level read of what a
representation or grounding choice fixes and what it leaves untouched.

\subsection{Cross-Task Transfer of the Taxonomy: StackCube-v1}
\label{sec:res-stack}
\begin{figure}[t]
  \centering
  \includegraphics[width=\linewidth]{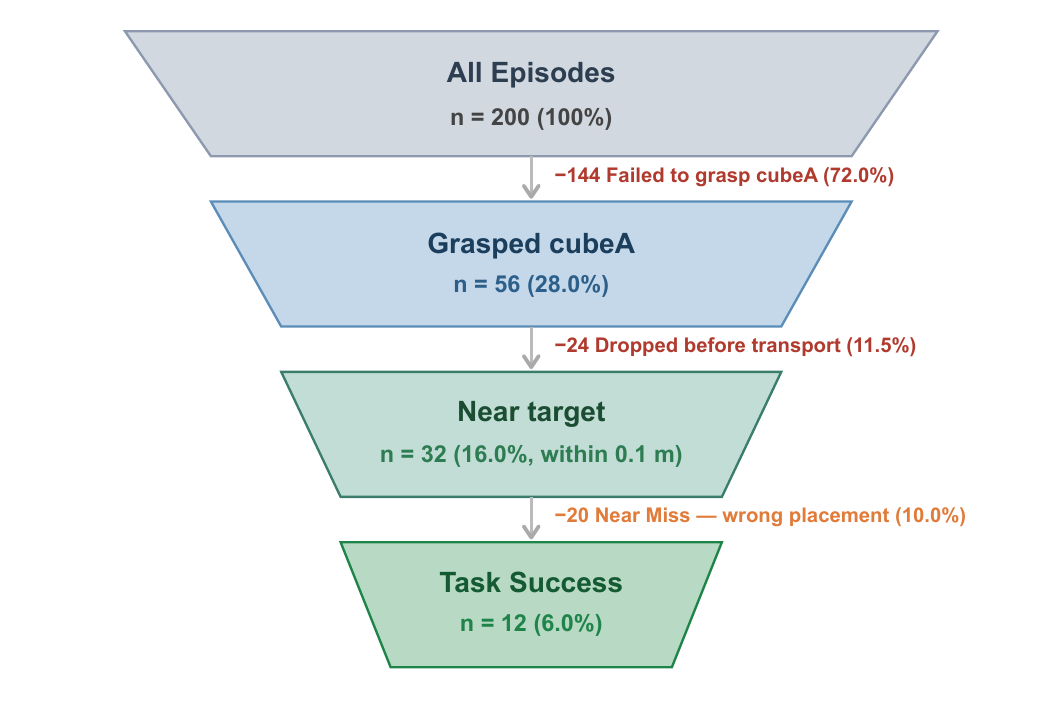}
  \caption{StackCube-v1 sub-task funnel (single 200-episode run).
    The dominant loss is at the grasp stage ($72\%$ No-Grasp), caused by
    arm-induced occlusion. This is a different bottleneck than PickCube's, and
    the same kinematic taxonomy applies unchanged.}
  \label{fig:funnel}
\end{figure}

To test whether the diagnostic carries over to another task, we train the 
same architecture on \textbf{StackCube-v1} and apply the same kinematic classifier without any change.
The policy reaches $5.3\!\pm\!2.2\%$ SR, well below the PickCube result. 
The taxonomy points straight at the bottleneck (Fig.~\ref{fig:funnel}):
No-Grasp accounts for $72\%$ ($144$ of $200$) of episodes, against $7.5\%$
($15$ of $200$) on PickCube under the pure-visual condition. The cause is
occlusion. During the close approach the arm hides the red cube from the base
camera, and our frozen SPOT has no temporal memory of slots, so it cannot
recover the cube once hidden. This is the opposite of the PickCube profile,
where Near Miss was the main failure and No-Grasp was stable. \textbf{The
taxonomy shows that the two tasks need different fixes}: StackCube needs temporal
slot consistency, not more spatial grounding. Binary success rate gives no such
guidance. The taxonomy works as a diagnostic that transfers across tasks.

\section{Discussion}
\label{sec:discussion}

\paragraph{Structure, not capacity.}
The results point to one statement: a small structural bottleneck on top of
frozen features generalizes where added parameters do not. Object-centric
grouping beats dense features by $22.4\%$ with the same trainable policy
(Sec.~\ref{sec:res-repr}), while cross-attention and longer temporal windows do
not improve transfer (Sec.~\ref{sec:res-fusion}). Slot Attention acts as a
capacity-limited, competition-based bottleneck: it caps how much the
representation can carry and pushes that capacity onto object-level regions.
Since the encoder is frozen, the only lever for an inductive bias is the
structure placed after the foundation model, not its weights.

\paragraph{One pattern across settings.}
This failure-and-fix shape is not unique to manipulation. In ViT
classification~\cite{shi_last-vit_2026}, in cross-domain 3D
perception~\cite{zhang_utonia_2026}, and in gaze
regression~\cite{qin_unigaze_2025}, a foundation model under global or
invariance-oriented supervision finds a low-cost cue that holds in-distribution
and breaks under shift, and the reported fix is structural rather than added
capacity. We show the pattern in one domain. We offer the parallel across
settings as context, not as a claim that it always holds.

\paragraph{Boundary of frozen slots.}
Three limitations bound the conclusions. First, SPOT has no temporal memory of
slots, so it cannot recover an object that becomes occluded mid-trajectory. This
is the direct cause of the StackCube collapse. Second, the fusion comparison was
run only without a goal token; we do not know whether concatenation is still
better once a reliable spatial anchor is present. Third, the representation
comparison uses approximate calibration. We show in
Sec.~\ref{sec:res-grounding} that exact calibration adds $+3.7\%$, so the
matched comparison is conservative but not calibration-free. Each limitation
points to a next step: temporal slot consistency for occlusion, a fusion
comparison with grounding present, and goal grounding derived from slot masks so
that camera extrinsics are no longer needed.

\section{Conclusion}
\label{sec:conclusion}
We presented a controlled, task-level study of visual representation types for
visuomotor imitation, evaluated on object and goal placements not seen in
training. Object-centric slot representations beat dense global and patch
features by $22.4\%$ under matched conditions, which shows that structure
matters more than token count. With an explicit 2D spatial goal, which is itself
derived from the ground-truth target, the full system reaches $68.7\%$, close to
the privileged 3D oracle ($71.7\%$). All of this uses a frozen encoder, a single
11\,GB GPU, and no end-to-end fine-tuning. A fusion ablation shows that simple
concatenation generalizes as well as cross-attention while overfitting less. An
automated kinematic failure taxonomy separates spatial-precision failures from
object-tracking failures, and transfers to StackCube-v1, where it points to
occlusion as the main bottleneck. For a practitioner on a small budget: prefer
object-centric slots to global features, use token concatenation rather than a
parametric fusion module, and give the policy an explicit spatial goal anchor.
Future work includes goal grounding from slot masks, temporal slot consistency
for occlusion, and sim-to-real transfer on a physical Franka Panda.

\section*{Acknowledgments}
This work builds on the first author's M.Sc.\ thesis carried out at the
Intelligent Autonomous Systems group, TU Darmstadt, and the LIRIS Lab,
École Centrale de Lyon.

\bibliographystyle{plainnat}
\bibliography{references}

\begin{thebibliography}{18}
\providecommand{\natexlab}[1]{#1}
\providecommand{\url}[1]{\texttt{#1}}
\expandafter\ifx\csname urlstyle\endcsname\relax
  \providecommand{\doi}[1]{doi: #1}\else
  \providecommand{\doi}{doi: \begingroup \urlstyle{rm}\Url}\fi

\bibitem[Cadena et~al.(2016)Cadena, Carlone, Carrillo, Latif, Scaramuzza,
  Neira, Reid, and Leonard]{cadena_slam_2016}
Cesar Cadena, Luca Carlone, Henry Carrillo, Yasir Latif, Davide Scaramuzza,
  Jos{\'e} Neira, Ian Reid, and John~J. Leonard.
\newblock Past, present, and future of simultaneous localization and mapping:
  Toward the robust-perception age.
\newblock \emph{IEEE Transactions on Robotics}, 32\penalty0 (6):\penalty0
  1309--1332, 2016.

\bibitem[Caron et~al.(2021)Caron, Touvron, Misra, J{\'e}gou, Mairal,
  Bojanowski, and Joulin]{caron_emerging_2021}
Mathilde Caron, Hugo Touvron, Ishan Misra, Herv{\'e} J{\'e}gou, Julien Mairal,
  Piotr Bojanowski, and Armand Joulin.
\newblock Emerging properties in self-supervised vision transformers.
\newblock In \emph{Proceedings of the IEEE/CVF International Conference on
  Computer Vision (ICCV)}, 2021.

\bibitem[Chapin et~al.(2026)Chapin, Machado, Dellandr{\'e}a, and
  Chen]{chapin_spotlighting_2026}
Alexandre Chapin, Bruno Machado, Emmanuel Dellandr{\'e}a, and Liming Chen.
\newblock Spotlighting task-relevant features: Object-centric representations
  for better generalization in robotic manipulation.
\newblock arXiv:2601.21416, 2026.

\bibitem[Chi et~al.(2023)Chi, Feng, Du, Xu, Cousineau, Burchfiel, and
  Song]{chi_diffusion_2023}
Cheng Chi, Siyuan Feng, Yilun Du, Zhenjia Xu, Eric Cousineau, Benjamin
  Burchfiel, and Shuran Song.
\newblock Diffusion policy: Visuomotor policy learning via action diffusion.
\newblock In \emph{Robotics: Science and Systems (RSS)}, 2023.

\bibitem[Elsayed et~al.(2022)Elsayed, Mahendran, van Steenkiste, Greff,
  Heigold, and Kipf]{elsayed_savi_2022}
Gamaleldin~F. Elsayed, Aravindh Mahendran, Sjoerd van Steenkiste, Klaus Greff,
  Georg Heigold, and Thomas Kipf.
\newblock {SAVi++}: Towards end-to-end object-centric learning from real-world
  videos.
\newblock In \emph{Advances in Neural Information Processing Systems
  (NeurIPS)}, 2022.

\bibitem[Kakogeorgiou et~al.(2024)Kakogeorgiou, Gidaris, Karantzalos, and
  Komodakis]{kakogeorgiou_spot_2024}
Ioannis Kakogeorgiou, Spyros Gidaris, Konstantinos Karantzalos, and Nikos
  Komodakis.
\newblock {SPOT}: Self-training with patch-order permutation for object-centric
  learning with autoregressive transformers.
\newblock In \emph{Proceedings of the IEEE/CVF Conference on Computer Vision
  and Pattern Recognition (CVPR)}, 2024.

\bibitem[Kipf et~al.(2022)Kipf, Elsayed, Mahendran, Stone, Sabour, Heigold,
  Jonschkowski, Dosovitskiy, and Greff]{kipf_conditional_2022}
Thomas Kipf, Gamaleldin~F. Elsayed, Aravindh Mahendran, Austin Stone, Sara
  Sabour, Georg Heigold, Rico Jonschkowski, Alexey Dosovitskiy, and Klaus
  Greff.
\newblock Conditional object-centric learning from video.
\newblock In \emph{International Conference on Learning Representations
  (ICLR)}, 2022.

\bibitem[Lee et~al.(2024)Lee, Wang, Etukuru, Kim, Shafiullah, and
  Pinto]{lee_vqbet_2024}
Seungjae Lee, Yibin Wang, Haritheja Etukuru, H.~Jin Kim, Nur Muhammad~Mahi
  Shafiullah, and Lerrel Pinto.
\newblock Behavior generation with latent actions.
\newblock In \emph{International Conference on Machine Learning (ICML)}, 2024.

\bibitem[Locatello et~al.(2020)Locatello, Weissenborn, Unterthiner, Mahendran,
  Heigold, Uszkoreit, Dosovitskiy, and Kipf]{locatello_object-centric_2020}
Francesco Locatello, Dirk Weissenborn, Thomas Unterthiner, Aravindh Mahendran,
  Georg Heigold, Jakob Uszkoreit, Alexey Dosovitskiy, and Thomas Kipf.
\newblock Object-centric learning with slot attention.
\newblock In \emph{Advances in Neural Information Processing Systems
  (NeurIPS)}, volume~33, 2020.

\bibitem[Nair et~al.(2022)Nair, Rajeswaran, Kumar, Finn, and
  Gupta]{nair_r3m_2022}
Suraj Nair, Aravind Rajeswaran, Vikash Kumar, Chelsea Finn, and Abhinav Gupta.
\newblock {R3M}: A universal visual representation for robot manipulation.
\newblock \emph{arXiv preprint arXiv:2203.12601}, 2022.

\bibitem[Qin et~al.(2025)Qin, Zhang, and Sugano]{qin_unigaze_2025}
Jiawei Qin, Xucong Zhang, and Yusuke Sugano.
\newblock {UniGaze}: Towards universal gaze estimation via large-scale
  pre-training.
\newblock In \emph{IEEE Winter Conference on Applications of Computer Vision
  (WACV)}, 2025.

\bibitem[Radosavovic et~al.(2022)Radosavovic, Xiao, James, Abbeel, Malik, and
  Darrell]{radosavovic_real-world_2022}
Ilija Radosavovic, Tete Xiao, Stephen James, Pieter Abbeel, Jitendra Malik, and
  Trevor Darrell.
\newblock Real-world robot learning with masked visual pre-training.
\newblock In \emph{Conference on Robot Learning (CoRL)}, 2022.

\bibitem[Seitzer et~al.(2023)Seitzer, Horn, Zadaianchuk, Zietlow, Xiao,
  Simon-Gabriel, He, Zhang, Sch{\"o}lkopf, Brox, and
  Engelcke]{seitzer_bridging_2023}
Maximilian Seitzer, Max Horn, Andrii Zadaianchuk, Dominik Zietlow, Tianjun
  Xiao, Carl-Johann Simon-Gabriel, Tong He, Zheng Zhang, Bernhard
  Sch{\"o}lkopf, Thomas Brox, and Martin Engelcke.
\newblock Bridging the gap to real-world object-centric learning.
\newblock In \emph{International Conference on Learning Representations
  (ICLR)}, 2023.

\bibitem[Shi et~al.(2026)Shi, Yu, and Yang]{shi_last-vit_2026}
Cheng Shi, Yizhou Yu, and Sibei Yang.
\newblock Vision transformers need more than registers.
\newblock In \emph{Proceedings of the IEEE/CVF Conference on Computer Vision
  and Pattern Recognition (CVPR)}, 2026.

\bibitem[Tao et~al.(2025)Tao, Xiang, Shukla, Qin, Hinrichsen, Yuan, Bao, Lin,
  Liu, kai Chan, Gao, Li, Mu, Xiao, Gurha, Rajesh, Choi, Chen, Huang, Calandra,
  Chen, Luo, and Su]{tao_maniskill3_2025}
Stone Tao, Fanbo Xiang, Arth Shukla, Yuzhe Qin, Xander Hinrichsen, Xiaodi Yuan,
  Chen Bao, Xinsong Lin, Yulin Liu, Tse kai Chan, Yuan Gao, Xuanlin Li,
  Tongzhou Mu, Nan Xiao, Arnav Gurha, Viswesh~Nagaswamy Rajesh, Yong~Woo Choi,
  Yen-Ru Chen, Zhiao Huang, Roberto Calandra, Rui Chen, Shan Luo, and Hao Su.
\newblock {ManiSkill3}: {GPU} parallelized robotics simulation and rendering
  for generalizable embodied {AI}.
\newblock \emph{Robotics: Science and Systems}, 2025.

\bibitem[Wu et~al.(2023)Wu, Hu, Lu, Gilitschenski, and
  Garg]{wu_slotdiffusion_2023}
Ziyi Wu, Jingyu Hu, Wuyue Lu, Igor Gilitschenski, and Animesh Garg.
\newblock {SlotDiffusion}: Object-centric generative modeling with diffusion
  models.
\newblock In \emph{Advances in Neural Information Processing Systems
  (NeurIPS)}, 2023.

\bibitem[Zhang et~al.(2026)Zhang, Wu, Yang, Fan, Li, Zhang, Huang, Wang, and
  Zhao]{zhang_utonia_2026}
Yujia Zhang, Xiaoyang Wu, Yunhan Yang, Xianzhe Fan, Han Li, Yuechen Zhang,
  Zehao Huang, Naiyan Wang, and Hengshuang Zhao.
\newblock {Utonia}: Toward one encoder for all point clouds.
\newblock In \emph{International Conference on Machine Learning (ICML)}, 2026.

\bibitem[Zhao et~al.(2023)Zhao, Kumar, Levine, and Finn]{zhao_learning_2023}
Tony~Z. Zhao, Vikash Kumar, Sergey Levine, and Chelsea Finn.
\newblock Learning fine-grained bimanual manipulation with low-cost hardware.
\newblock In \emph{Robotics: Science and Systems (RSS)}, 2023.

\end{thebibliography}

\end{document}